\pdfoutput=1

\documentclass[11pt]{article}

\usepackage[final]{acl}
\usepackage{inconsolata}
\usepackage{xcolor}
\usepackage{times}
\usepackage{latexsym}
\usepackage[T1]{fontenc}
\usepackage[utf8]{inputenc}
\usepackage{amsmath}
\usepackage{mathtools}
\usepackage{todonotes}
\usepackage{comment}
\usepackage{rotating}
\usepackage{adjustbox}
\usepackage{changepage}

\newcommand{\dblvert}[1]{\left\| #1 \right\|}

\usepackage{multicol}
\usepackage{multirow}
\usepackage{booktabs}
\usepackage{algorithmic}
\usepackage{algorithm}
\usepackage{amsmath}

\usepackage{graphicx}

\title{Rethinking ASTE: A Minimalist Tagging Scheme Alongside Contrastive Learning}

\author{Qiao Sun$^{1,2}$ \quad Liujia Yang$^{2,3}$ \quad Minghao Ma$^{1}$ \quad Nanyang Ye$^{3}$ \quad Qinying Gu$^{2}$ \\
$^{1}$Fudan University \\
$^{2}$Shanghai AI Lab \\
$^{3}$Shanghai Jiao Tong University \\
\texttt{qiaosun22@m.fudan.edu.cn} \quad \texttt{20307130024@fudan.edu.cn} \\
\texttt{yangliujia1008@sjtu.edu.cn} \quad \texttt{ynylincolncam@gmail.com} \\
\texttt{qinying.gu0220@gmail.com}
}

\begin{document}
\maketitle

\begin{abstract}
Aspect Sentiment Triplet Extraction (ASTE) is a burgeoning subtask of fine-grained sentiment analysis, aiming to extract structured sentiment triplets from unstructured textual data. Existing approaches to ASTE often complicate the task with additional structures or external data. In this research, we propose a novel tagging scheme and employ a contrastive learning approach to mitigate these challenges. The proposed approach demonstrates comparable or superior performance in comparison to state-of-the-art techniques, while featuring a more compact design and reduced computational overhead. Notably, even in the era of Large Language Models (LLMs), our method exhibits superior efficacy compared to GPT 3.5 and GPT 4 in a few-shot learning scenarios. This study also provides valuable insights for the advancement of ASTE techniques within the paradigm of LLMs.
\end{abstract}

\section{Introduction}
\begin{figure}[!t]
    \begin{center}
    \includegraphics[scale=0.475]{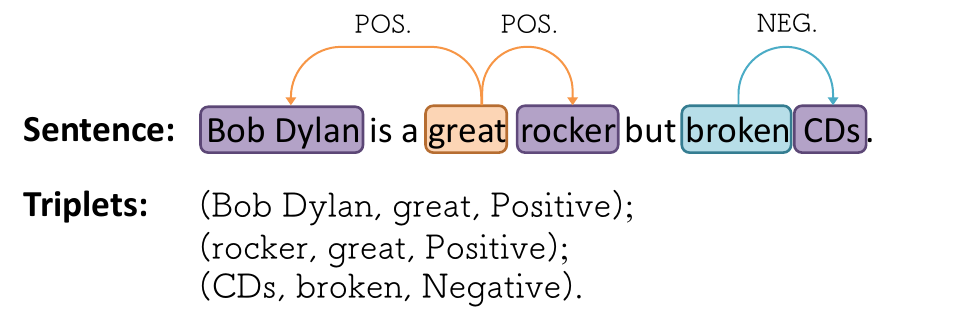} 
    \caption{
    An example for the ASTE task illustrating \texttt{Aspect} terms in purple, \texttt{Opinion} terms with \texttt{Negative} sentiment in blue, and \texttt{Opinion} terms with \texttt{Positive} sentiment in orange.
    }
    \label{fig.example}
    \end{center}
\end{figure}

Aspect Sentiment Triplet Extraction (ASTE) is an emerging fine-grained\footnote{Generally, sentiment analysis can be based on three levels, namely, document-based, sentence-based, and aspect-based \citep{jing2021seeking}.} sentiment analysis task \cite{pontiki2014novel, pontiki2015semeval, pontiki2016semeval} aimed at identifying and extracting structured sentiment triplets \cite{peng2020knowing}, defined as \texttt{(Aspect, Opinion, Sentiment)}, from unstructured text. 
Specifically, an \texttt{Aspect} term refers to the subject of discussion, an \texttt{Opinion} term provides a qualitative assessment of the \texttt{Aspect}, and \texttt{Sentiment} denotes the overall sentiment polarity, typically taken from a three-level scale \texttt{(Positive, Neutral, Negative)}. 
For instance, consider the sentence: ``The battery life is good, but the camera is mediocre.''
The ground truth result is \texttt{\{(battery life, good, Positive), (camera, mediocre, Neutral)\}}. Figure \ref{fig.example} illustrates another example. 
Recent methods \cite{wu2020clear, jing2021seeking, chen2022enhanced, zhang2022boundary, liang2023stage}  commonly utilize Pretrained Language Models (PLMs) to encode input text. The powerful representational capacity of PLMs has greatly advanced the performance in this field, yet there is a tendency to employ complex classification head designs and leverage information enhancement techniques to achieve marginal performance improvements. 

In this work, we attribute the current challenges in ASTE to two main factors:
1) \textit{the longstanding overlook of the conical embedding distribution problem} and 
2) \textit{imprudent tagging scheme design}. 
We critically examine the conventional 2D tagging method, commonly known as the table-filling approach, to reassess the efficacy of tagging schemes. Highlighting the critical role of tagging scheme optimization, we delve into what constitutes an ideal scheme for ASTE.
We analyze the advantages of the full matrix approach over the half matrix approach and decompose the labels in the full matrix into 1) \textit{location} and 2) \textit{classification}. Thanks to this, we come up with a new tagging scheme with a minimum number of labels to effectively reduce the complexity of training and inference. 
Moreover, this tagging scheme can be well aligned with our novel contrastive learning mechanism. 
To the best of our knowledge, this is the first formal analysis of the tagging scheme to guide a rational design and the first attempt to adopt token-level contrastive learning to improve the PLM representations' distribution and facilitate the learning process. 




The contributions of this work can be summarized as follows:
\begin{itemize}
    \item We offer the first critical evaluation of the 2D tagging scheme, particularly focusing on the table-filling method. This analysis pioneers in providing a structured framework for the rational design of tagging schemes.
    \item We introduce a simplified tagging scheme with the least number of label categories to date, integrating a novel token-level contrastive learning approach to enhance PLM representation distribution.
    \item Our study addresses ASTE challenges in the context of LLMs, developing a tailored in-context learning strategy. Through evaluations on GPT 3.5-Turbo and GPT 4, we establish our method's superior efficiency and effectiveness.
\end{itemize}


\section{Method}
\begin{figure*}[!ht]
\begin{center}
\includegraphics[scale=0.375]{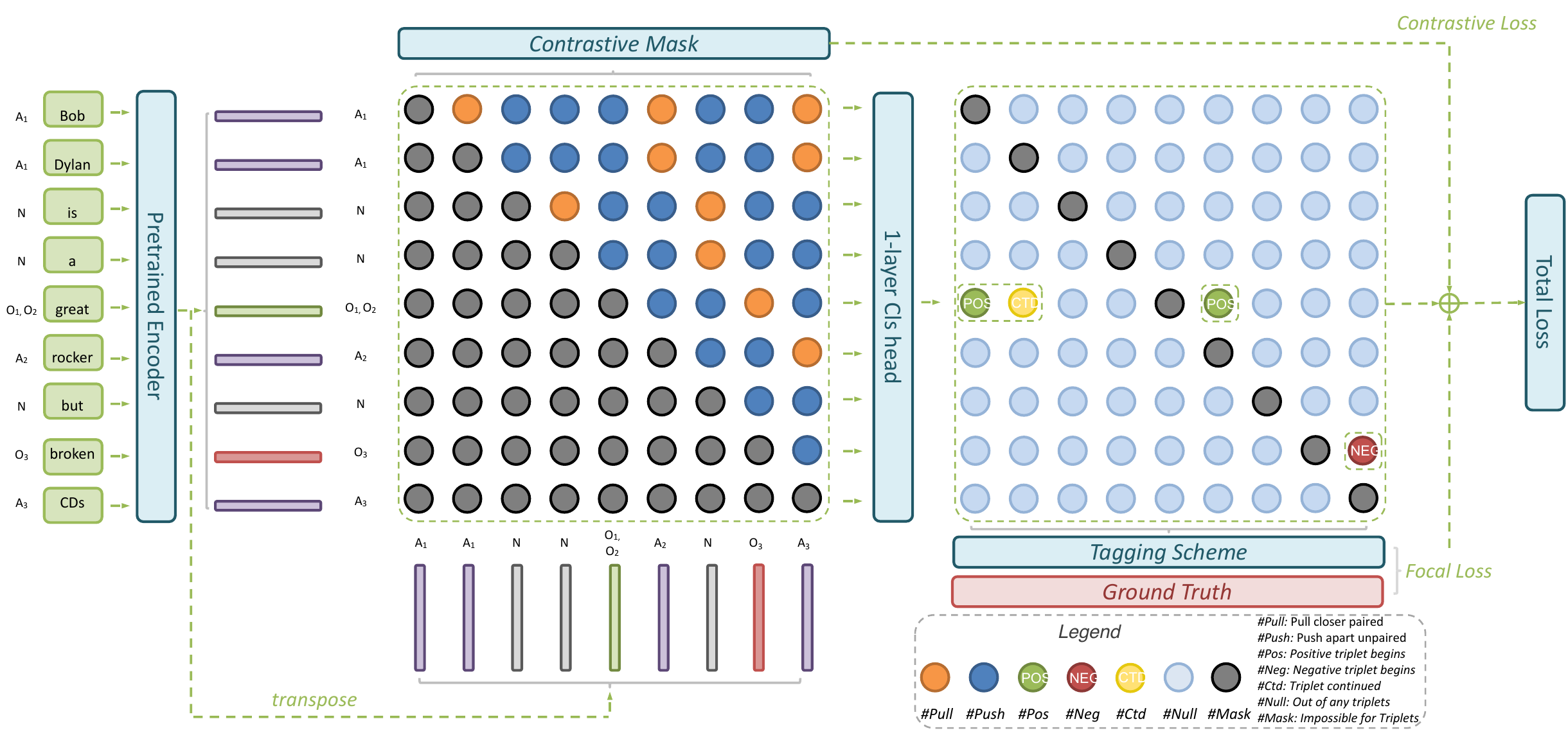} 
\caption{Schematic diagram of our proposed framework which contains: 1) a contrastive learning mechanism that aligns \texttt{Aspect} and \texttt{Opinion}; and 2) a tagging scheme encompassing 5 label classes: \texttt{NULL}, \texttt{CTD}, \texttt{POS}, \texttt{NEU}, and \texttt{NEG}.}
\label{fig.framework}
\end{center}
\end{figure*}

Figure \ref{fig.framework} presents our framework. Essentially, our method contributes by two aspects: 1) a \textit{contrastive-learning-enhanced PLM encoder} and 2) a \textit{minimalist tagging scheme}.
\textbf{Appendix} Algorithm \ref{alg1} further delivers a pseudo-code for the training process of our proposed framework. 

\subsection{The Contrastive-learning-enhanced PLM Encoder}
Note that a representation distribution satisfying \textit{alignment} and \textit{uniformity} is linearly separable \cite{wu2023hallucination} and facilitates the classification. Thereby, contrastive learning boosts represent learning by improving the \textit{alignment} and \textit{uniformity} of the representations \cite{wu2023hallucination}. However, recent investigations indicate that representation distributions in pretrained models often diverge from these expectations. 
\citeauthor{liang2021learning} (\citeyear{liang2021learning}) computed similarities for randomly sampled word pairs, revealing that word embeddings in an Euclidean space cluster within a confined cone, rather than uniformly distributed.

The motivation to adopt contrastive learning is hence to improve the distribution of representations output by PLMs. The core idea lies in that, after fine-tuning on a specific task, the PLM encoder should embed words with similar roles to distribute closer and drive the different ones to be farther.

Given the input sentence $\mathcal{S}$ and pretrained model \texttt{Encoder}
, it outputs the hidden word embeddings $\mathcal{H}_{|\mathcal{H}|}$:
    \begin{equation}
\mathcal{H}_{|\mathcal{H}|}=\texttt{Encoder}(\mathcal{S}_{|\mathcal{S}|}), 
    \end{equation}
\noindent where $\mathcal{H}_{|\mathcal{H}|}=(h_1, h_2, ..., h_{|\mathcal{H}|})$. Note that, $|\mathcal{H}|$ represents the number of tokens $\mathcal{H}$, which is not necessarily equal the number of words $\mathcal{S}$.


Inspired by \citep{schroff2015facenet}, 
we 1) take the Euclidean distance as a negative metric on the similarity and 2) 
introduce a margin $d$ to enforce the gap 
between two similar hidden word representations, that is, stop pulling $\mathcal{H}_i$ and $\mathcal{H}_j$
closer when there is $\dblvert{\mathcal{H}_i - \mathcal{H}_j}^2 \leq d$ (avoiding similar representations from squeezing too much with each other).

So, the metric of similarity is 

\begin{equation}
\begin{aligned}
\mathrm{Sim}_{i, j}=\texttt{Sim}(\mathcal{H}_i, \mathcal{H}_j)=-\dblvert{\mathcal{H}_i - \mathcal{H}_j}^2
\end{aligned}
\end{equation}

Hence, the similarities $\mathrm{Sim}_{i, j}, 
i, j \in \{1, 2, ..., |\mathcal{H}|\}$ forms a similarity matrix $\mathbf{Sim}_{|\mathcal{H}| \times |\mathcal{H}|}$. 

To make the representations closer among tokens within the same class and farther between that of different classes, while keeping a margin of $d$, we can simply maximize $\mathrm{Sim}_{i, j}$ if $\mathcal{H}_i$ and $\mathcal{H}_j$ shares the same class and else minimize $\mathrm{Sim}_{i, j}$.


Defining a matrix $\mathbf{M_{|\mathcal{H}|\times|\mathcal{H}|}}$, it controls whether to pull (closer) or push (farther) between the hidden representation of words, that is, the ``Contrastive Mask'' used to calculate the contrastive loss. In Figure \ref{fig.framework}, it is the left-side strict upper-triangle matrix in blue, orange and grey\footnote{
The meaning of colors: The blue cell means that on this cell the row representation is in a different class from the column representation, and the orange cell means that with same class.
}. 

\begin{equation}
\small
\begin{aligned}
  &\mathcal{L}_{contrastive}\\
  =&\sum_{(\mathcal{H}_i, \mathcal{H}_j)^+} \max\{\dblvert{\mathcal{H}_i - \mathcal{H}_j}^2,\ d\}\\
  -&\sum_{(\mathcal{H}_i, \mathcal{H}_j)^-} \min\{\dblvert{\mathcal{H}_i - \mathcal{H}_j}^2,\ d\}\\
  =&\sum_{i=1}^{|\mathcal{H}|}\sum_{j=i+1}^{|\mathcal{H}|}\max\{\texttt{Sim}(\mathcal{H}_i, \mathcal{H}_j),\ d\}\cdot \mathbf{M}_{i, j}.\\
  =&\sum_{i=1}^{|\mathcal{H}|}\sum_{j=1}^{|\mathcal{H}|}\big(\max\{\mathbf{Sim_{|\mathcal{H}|\times|\mathcal{H}|}},\ d\} \circ \mathbf{M_{|\mathcal{H}|\times|\mathcal{H}|}}\big)_{i, j}
\end{aligned}
\end{equation}

\noindent where $\circ$ is a notation for the Hadamard product \cite{horn1990hadamard}. 

\begin{figure*}[!ht]
\begin{center}
\includegraphics[scale=0.72]{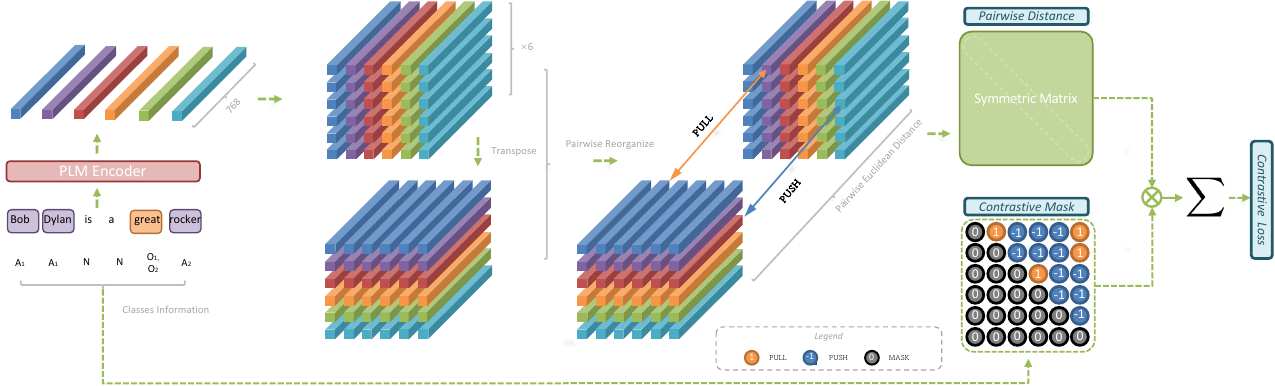} 
\caption{A more detailed illustration of our contrastive learning mechanism.}
\label{fig.clmatrix}
\end{center}
\end{figure*}

Figure \ref{fig.clmatrix} depicts the workflow of our contrastive learning strategy and showcases the detail of $\mathbf{M_{|\mathcal{H}|\times|\mathcal{H}|}}$. 
Note that for the purpose of contrastive learning, the order of each pair of words is redundant, so using the strict upper matrix is enough and we mask the lower triangle. 


Noteworthily, 
a PLM encoder, like BERT, encodes the same word input to different vector representations when the context is different. For example, suppose two sentences, 1) ``I really just like the old school plug-in ones'', and 2) ``The old school is beautiful''. Explicitly the ``school'' should be labelled as \texttt{Aspect} in the former case whilst as \texttt{Opinion} in the latter. 
For these cases, a well-learned PLM encoder will encode the unique word ``school'' into different representations according to the context. 
This characteristic allows for the direct application of contrastive learning to the hidden word representations without worrying about contradiction. 

\subsection{The Minimalist Tagging Scheme}


Rethinking the 2D tagging scheme:

\noindent \textbf{Lemma 1}. Specific to the ASTE task, when we take it as a 2D-labeling problem, we are to 1) find a set of tagging strategies to establish a 1-1 map between each triplet and its corresponding tagging matrix. See the proof in \textbf{Appendix} Proof 1. 
\label{lemma1}

\noindent \textbf{Lemma 2}. In a 2D-tagging for ASTE, at least three basic goals must be met: 
1) correctly identifying the \texttt{(Aspect, Opinion)} pairs, 
2) correctly classifying the sentiment polarity of the pair based on the context, and 
3) avoiding boundary errors, such as \textit{overlapping}\footnote{It occurs when one single word belongs to multiple classes in different triplets. }, \textit{confusion}\footnote{It occurs when there is a lack of location restrictions so that multiple neighbored candidates can not be uniquely distinguished. }, 
and \textit{conflict}\footnote{It occurs when one single word is composed of multiple tokens, and the predict gives predictions that are not aligned with the word span. }. See the proof in \textbf{Appendix} Proof 2.


\noindent \textbf{Theorem 1}. From insight of the above lemmas, it can be concluded that using \textbf{enough} (that is, following the 1-1 map properties in Lemma 1, as well as avoiding the issues in Lemma 2) labels will make it a theoretically ensured tagging scheme.

\noindent \textbf{Assumption 1}. Ceteris paribus, for a specific classification neural network, the \textbf{fewer} the number of target categories, the easier it is for the network to learn. This is a empirical and heuristic assumption, for the reasonable consideration of \textit{Simplification of Decision Boundaries} \cite{Hinton2006} and \textit{Enhancement of Training Efficiency} (less parameters). 

Combining Theorem 1 and Assumption 1, \textbf{fewer} yet \textbf{enough} labels can be heuristically better solution with theoretical guarantee.   



\begin{figure*}[!ht]
\begin{center}
\includegraphics[scale=0.38]{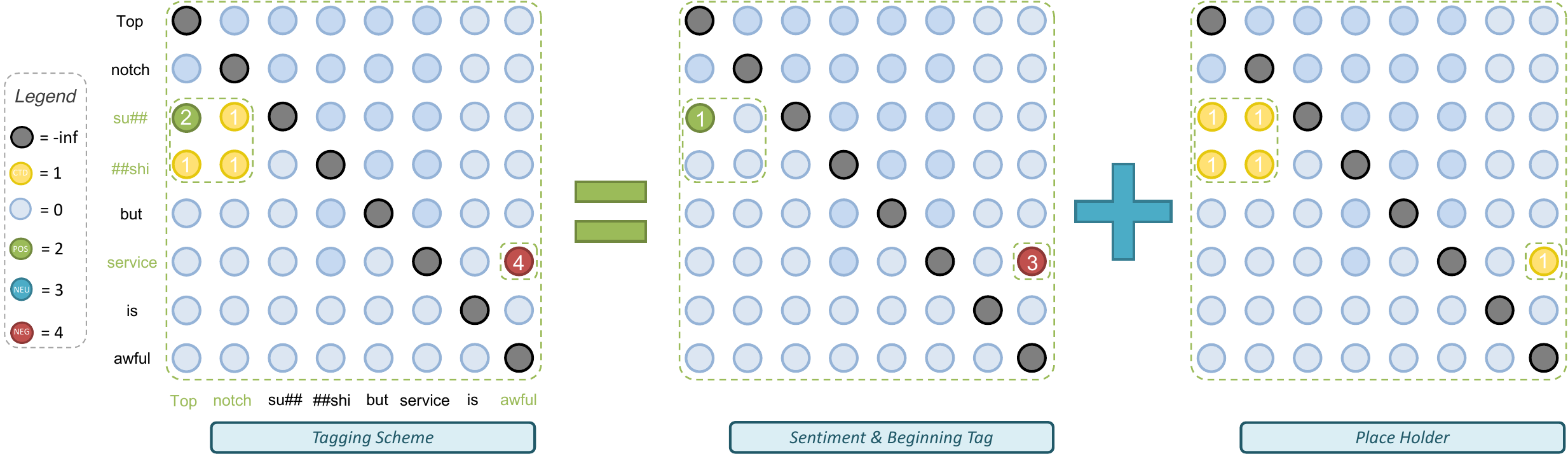} 
\caption{Decomposition of the tagging scheme into two components: 1) a beginning mark matrix with sentiment labels; and 2) a placeholder matrix denoting regions of triplets with ``1''s and default regions with ``0''s. Remember that each row is taken as candidates for an \texttt{Aspect} and each column is taken as candidates for an \texttt{Opinion}. Naturally, each cell in the square matrix can be seen as an ordered pair for a unique candidate of \texttt{<Aspect, Opinion>}. When we simply sum the two components up, we have the left-hand tagging scheme in Figure \ref{fig.tagging}, where the ``Sentiment \& Beginning Tag'' is like a trigger (just like you click your mouse), and the ``Place Holder'' is like a ``continued shift'' (continue to hold and drag the mouse to the downright).}
\label{fig.tagging}
\end{center}
\end{figure*}

With the above knowledge, our tagging scheme employs a full matrix (illustrated as Figure \ref{fig.tagging}) so that rectangular occupations in its cells indicate respective triplets, where each of the rectangles' row indices correspond to the relative \texttt{Aspect} term and the column indices correspond to the \texttt{Opinion}. Hereafter, this kind of labels can be taken as a set of ``place holder'', which is obviously a 1-1 map meeting Lemma 1. 


To further satisfy Lemma 2, we introduce another kind of labels, ``sentiment \& beginning tag''. This set of labels specializes in recognizing the top-left corner of a ``shadowed'' area. Meanwhile, it takes a value from the sentiment polarity, i.e. \texttt{Positive, Neutral, Negative}. This tagging is crucial to both \textit{identify the beginning of an triplet} and \textit{label the sentiment polarity}. 

Figure \ref{fig.tagging} shows a comprehensive case of our tagging scheme, in which the left matrix is an appearance of our tagging scheme, and it can be decomposed into two separate components. The middle matrix is the first component,  which takes only one tag to locate the up-left beginning of an area, and the second component simply predicts a binary classification to figure out the full area.

Note that, this design benefits the tagging scheme's decode process. By scanning across the matrix, we only start an examination function when triggered by a beginning label like this, and then search by row and column until it meets any label except a ``continued'' (``CTD''), which satisfies Lemma 2.

Thanks to the above design, the ASTE task is well addressed with \textbf{ONLY} employing a simple classification head as follows:
\begin{equation}
\begin{aligned}
&H_1 = \texttt{LinearClsHead}(\mathcal{H}_{contrasted})\\
&H_3 = \texttt{LayerNorm}(H_2)\\
&\mathbf{Tag}_{pred} = \texttt{GELU}(H_3)
\end{aligned}
\end{equation}

\section{Experiments}
\subsection{Implementation Details}

All experiments were conducted on a single RTX 2080 Ti. 
The best model weight on the development set is saved and then evaluated on the test set. For the PLM encoder, the pretrained weights \texttt{bert\_base\_uncased} and \texttt{roberta\_base} are downloaded from \citep{wolf2020transformers}. GPT 3.5-Turbo and GPT 4 are implemented using OpenAI API \cite{openai_api}.
The learning rate is $1\times 10^{-5}$ for the PLM encoder, and $1\times 10^{-3}$ for the classification head. 


\subsection{Datasets}
We evaluate our method on two canonical ASTE datasets derived from the SemEval Challenges \cite{pontiki2014novel, pontiki2015semeval, pontiki2016semeval}. These datasets serve as benchmarks in the majority of Aspect-based Sentiment Analysis (ABSA) research. The first dataset, denoted as $\mathcal{D}_1$, is the Aspect-oriented Fine-grained Opinion Extraction (AFOE) dataset introduced by \cite{wu2020grid}. The second dataset, denoted as $\mathcal{D}_2$, is a refined version by \cite{xu2020position}, building upon the work of \cite{peng2020knowing}. Further details are provided in Table \ref{tab:data1}.

\subsection{Baselines}
We evaluate our method against various techniques including pipeline, sequence-labeling, seq2seq, table-filling and LLM-based approaches. Detailed descriptions for each method can be found in the \textbf{Appendix} Table \ref{baseline}.



\begin{table*}[htbp]
\centering
  \scalebox{0.68}{
    \begin{tabular}{cccccccccccccccc}
    \hline
    \toprule
    \multirow{2}[4]{*}{\textbf{Methods}} & \multicolumn{3}{c}{\textbf{14Res}} &       & \multicolumn{3}{c}{\textbf{14Lap}} &       & \multicolumn{3}{c}{\textbf{15Res}} &       & \multicolumn{3}{c}{\textbf{16Res}} \\
\cmidrule{2-4}\cmidrule{6-8}\cmidrule{10-12}\cmidrule{14-16}          & P     & R     & F1    &       & P     & R     & F1    &       & P     & R     & F1    &       & P     & R     & F1 \\
    \midrule
    \midrule
    \textbf{Pipeline} &       &       &       &       &       &       &       &       &       &       &       &       &       &       &  \\
    $\textrm{Two-stage}^\natural$\ \textrm{\cite{peng2020knowing}} & 43.24 & 63.66 & 51.46 &       & 37.38 & 50.38 & 42.87 &       & 48.07 & 57.51 & 52.32 &       & 46.96 & 64.24 & 54.21 \\
    $\textrm{Li-unified-R+PD}^\sharp$\ \textrm{\cite{peng2020knowing}} & 40.56 & 44.28 & 42.34 &       & 41.04 & 67.35 & 51.00 &       & 44.72 & 51.39 & 47.82 &       & 37.33 & 54.51 & 44.31 \\
    \midrule
    \textbf{Sequence-tagging} &       &       &       &       &       &       &       &       &       &       &       &       &       &       &  \\
    Span-BART \textrm{\cite{yan2021unified}}  & 65.52 & 64.99 & 65.25 &       & 61.41 & 56.19 & 58.69 &       & 59.14 & 59.38 & 59.26 &       & 66.60 & 68.68 & 67.62 \\
    JET \textrm{\cite{xu2020position}}  & 70.56 & 55.94 & 62.40 &       & 55.39 & 47.33 & 51.04 &       & 64.45 & 51.96 & 57.53 &       & 70.42 & 58.37 & 63.83 \\
    \midrule
    \textbf{Seq2seq} &       &       &       &       &       &       &       &       &       &       &       &       &       &       &  \\
    Dual-MRC \textrm{\cite{mao2021joint}} & 71.55 & 69.14 & 70.32 &       & 57.39 & 53.88 & 55.58 &       & 63.78 & 51.87 & 57.21 &       & 68.60 & 66.24 & 67.40 \\
    $\textrm{BMRC}^\dagger$ \textrm{\cite{chen2021bidirectional}} & 72.17 & 65.43 & 68.64 &       & 65.91 & 52.15 & 58.18 &       & 62.48 & 55.55 & 58.79 &       & 69.87 & 65.68 & 67.35 \\
    COM-MRC \textrm{\cite{zhai2022mrc}} & 75.46 & 68.91 & 72.01 &       & 62.35 & 58.16 & 60.17 &       & 68.35 & 61.24 & 64.53 &       & 71.55 & 71.59 & 71.57 \\ 
    Triple-MRC \textrm{\cite{zou2024multi}} & - & - & 72.45 &       & - & - & 60.72 &       & - & - & 62.86 &       & - & - & 68.65 \\
    
    \midrule
    \textbf{Table-filling} &       &       &       &       &       &       &       &       &       &       &       &       &       &       &  \\
    GTS \textrm{\cite{wu2020grid}}   & 67.76 & 67.29 & 67.50 &       & 57.82 & 51.32 & 54.36 &       & 62.59 & 57.94 & 60.15 &       & 66.08 & 66.91 & 67.93 \\
    Double-encoder \textrm{\cite{jing2021seeking}} & 67.95 & 71.23 & 69.55 &       & 62.12 & \underline{56.38} & 59.11 &       & 58.55 & 60.00 & 59.27 &       & 70.65 & 70.23 & 70.44 \\
    EMC-GCN \textrm{\cite{chen2022enhanced}} & 71.21 & 72.39 & 71.78 &       & 61.70 & 56.26 & 58.81 &       & 61.54 & 62.47 & 61.93 &       & 65.62 & 71.30 & 68.33 \\

    BDTF \textrm{\cite{zhang2022boundary}} & 75.53 & \underline{73.24} & \underline{74.35} &       & 68.94 & 55.97 & 61.74 &       & 68.76 & \underline{63.71} & \textbf{66.12} &       & 71.44 & \underline{73.13} & 72.27 \\
    STAGE-1D \textrm{\cite{liang2023stage}} & \textbf{79.54} & 68.47 & 73.58 &       & \underline{71.48} & 53.97 & 61.49 &       & 72.05 & 58.23 & 64.37 &       & \textbf{78.38} & 69.10  & \underline{73.45} \\
    STAGE-2D \textrm{\cite{liang2023stage}} & 78.51 & 69.3  & 73.61 &       & 70.56 & 55.16 & \underline{61.88} &       & \underline{72.33} & 58.93 & 64.94 &       & \underline{77.67} & 68.44 & 72.75 \\
    STAGE-3D \textrm{\cite{liang2023stage}} & \underline{78.58} & 69.58 & 73.76 &       & \textbf{71.98} & 53.86 & 61.58 &       & \textbf{73.63} & 57.9  & 64.79 &       & 76.67 & 70.12 & 73.24 \\
    DGCNAP \textrm{\cite{li2023dual}} & 72.90 & 68.69 & 70.72 &       & 62.02 & 53.79 & 57.57 &       & 62.23 & 60.21  & 61.19 &       & 69.75 & 69.44 & 69.58 \\
    \midrule

    \textbf{LLM-based} &       &       &       &       &       &       &       &       &       &       &       &       &       &       &  \\
    GPT 3.5 zero-shot  & 44.88     & 55.13     & 49.48 &       & 30.04     & 41.04     & 34.69 &       & 36.02     & 53.40     & 43.02 &       & 39.92     & 57.78     & 47.22 \\
    $\textrm{GPT 3.5 few-shots}$\  & 52.36     & 54.63     & 53.47 &       & 29.91     & 36.04     & 32.69 &       & 45.48     & 61.44     & 52.01 &       & 49.50     & 67.12     & 56.98 \\
    \textrm{GPT 4 zero-shot}  & 32.99 & 38.13 & 35.37 &       & 17.81 & 22.55 & 19.90 &       & 27.85 & 37.73 & 32.05 &       & 32.17 & 43.00 & 36.80 \\
    $\textrm{GPT 4 few-shots}$\  & 47.25 & 49.20 & 48.20 &       & 26.04 & 33.64 & 29.35 &       & 39.94 & 51.13 & 44.85 &       & 43.72 & 54.86 & 48.66 \\
    \midrule
    
    \textbf{Ours} &       &       &       &       &       &       &       &       &       &       &       &       &       &       &  \\
       ContrASTE   & 76.1  & \textbf{75.08} & \textbf{75.59} &       & 66.82 & \textbf{60.68} & \textbf{63.61}&       & 66.50 & \textbf{63.86} & \underline{65.15} &       & 75.52 & \textbf{74.14} & \textbf{74.83} \\
    \bottomrule
    \hline
    \end{tabular}%
    }
\caption{Experimental results on $\mathcal{D}_2$ \cite{xu2020position}. 
The best results are highlighted in bold,  while the second best results are underlined.
}

\label{tab-performance2}
\end{table*}

\subsection{Performance on ASTE Task}





We evaluate ASTE performance using the widely accepted \texttt{(Precision, Recall, F1)} metrics. 
The result on dataset $\mathcal{D}_2$ can be found in Table \ref{tab-performance2} and on $\mathcal{D}_1$ is presented in \textbf{Appendix} Table \ref{tab-performance1}. The best results are indicated in bold, while the second best results are underlined. Our proposed method consistently achieves state-of-the-art performance or ranks second across all evaluated cases.

Significantly, on dataset $\mathcal{D}_1$, proposed method achieves a substantial improvement of 3.08\% in F1 score on the 14Lap subset. This improvement is particularly noteworthy considering that the best score on this dataset is the lowest among all the datasets, showcasing our ability to effectively handle challenging instances. Moreover, on the 14Res subset, our F1 score surpasses 76.00+, which, to the best of our knowledge, is the highest reported performance.

Turning to dataset $\mathcal{D}_2$, our method outperforms all state-of-the-art approaches by more than 1 percentage point on the 14Res, 14Lap, and 16Res subsets. Only on the 16Res subset, the BDTF method \citep{zhang2022boundary} achieves a slightly better performance. 

On both datasets, our approach overwhelms GPTs' substantially,  indicating that our proposed method keeps advancement in the LLM era. \textbf{Appendix} Table \ref{gptExample} showcases more interesting facts of GPT's performance.


We highlight that we evaluate our method against LLM-based approaches in zero-shot and few-shots cases, which could be blamed for not adopting a full parameter fine-tuning. Nevertheless, our evaluation of GPT models on ASTE tasks stems from curiosity about the processing capabilities of large language models, which is believed to be a common interest in the ASTE community, where the full parameter fine-tuning may result in catastrophic forgetting \cite{lin2024mitigating} (which is unaffordable in a specific task like ASTE) and is left for future research efforts. 

\subsection{Performance on Other ABSA Tasks}
Our method can also effectively handle other ABSA subtasks, including Aspect Extraction (AE), Opinion Extraction (OE), and Aspect Opinion Pair Extraction (AOPE).
AE aims to extract all the \texttt{(Aspect)} terms, OE aims to extract all the \texttt{Opinion} terms, and AOPE aims to extract all the \texttt{(Aspect, Opinion)} pairs from raw text. 
The results for these tasks are presented in Table \ref{ABSA} in the \textbf{Appendix}, where our method consistently achieves best F1-scores across nearly all tasks.

\begin{table}[htbp]
  \centering
  \scalebox{0.52}{
    \begin{tabular}{lccccccccc}
    \hline
    \toprule
    \multirow{2}[4]{*}{\textbf{Models}} & \multicolumn{4}{c}{$\mathcal{D}_1$}        &       & \multicolumn{4}{c}{$\mathcal{D}_2$} \\
\cmidrule{2-5}\cmidrule{7-10}          & \textbf{14Res} & \textbf{14Lap} & \textbf{15Res} & \textbf{16Res} &       & \textbf{14Res} & \textbf{14Lap} & \textbf{15Res} & \textbf{16Res} \\
    \midrule
    \midrule
    \textbf{ContrASTE}  & 76.00    & 64.07  & 65.43 & 71.80 &       & 75.59 & 63.61 & 65.15 & 74.83 \\
    \midrule
        w/o. RoBERTa & 74.12 & 63.18 & 62.95 & 69.41 &       & 72.66 & 62.15 & 63.25 & 70.71 \\
    \ \ \ \ $\Delta$F\_1  & -1.88 & -0.89 & -2.48  & -2.39  &       & -2.93 & -1.46 & -1.90    & -4.12 \\
    \midrule
        \ \ \ \ w/o. contr & 72.61 & 61.94 & 58.14 & 68.16 &       & 71.72 & 61.49 & 58.11 & 68.03 \\
    \ \ \ \ $\Delta$F\_1  & -3.39 & -2.13 & -7.29 & -3.64 &       & -3.87 & -2.12 & -7.04 & -6.80 \\
    \midrule
        \ \ \ \ w/o. tag & 67.78 & 54.98 & 60.75 & 62.62 &       & 65.83 & 54.98 & 58.73 & 67.63 \\
    \ \ \ \ $\Delta$F\_1  & -8.22 & -9.09 & -4.68 & -9.18 &       & -9.76 & -8.63 & -6.42 & -7.20 \\
    \bottomrule
    \hline
    \end{tabular}%
    }
    \caption{Ablation study on 
    F1, 
    where ``w/o. RoBERTa'' denotes ``Replace RoBERTa with bert-base-uncased'', ``w/o. contr'' denotes without the contrastive learning mechanism, and ``w/o. tag'' denotes ``replace our tagging scheme with a baseline''.}
    \label{ablation}%
\end{table}%

\section{Analysis}

\subsection{Ablation Study}
 \textbf{Encoder}. 
In the ablation experiment part, by replacing RoBERTa by BERT, The results obtained declined slightly while still yield most other methods.

\noindent\textbf{Contrastive Learning}. First, we deactivating the contrastive mechanism in our method (denoting ``w/o. contr'') by setting the coefficient of the contrastive loss to 0. The results in Table \ref{ablation} illustrate a significant F1-score decrease of $0.22\sim 2.93$\%.


\noindent\textbf{Tagging Scheme}. Third, we substitute our proposed scheme with the conventional GTS tagging scheme \cite{wu2020grid}, which results in a significant performance decline (Table \ref{ablation}) by \textcolor{black}{$1.46\sim6.14$\%}. This indicates that the contrastive learning methods, within our framework, is of strong reliance on an appropriate tagging scheme. This reinforces the effectiveness of our straightforward yet impactful tagging scheme.

\subsection{Effect of Contrastive Learning}
In \textbf{Appendix} Figure \ref{fig.clrepre} an example is shown of how contrastive learning has improved the representation, where the left subplot is the tokens output by RoBERTa without contrastive learning, and the right one is with contrastive learning. Note that the Principal Component Analysis (PCA) \citep{mackiewicz1993principal} is adopted to reduce the dimensions of the vectors to 2. 

\begin{figure*}[ht]
\begin{center}
\includegraphics[scale=0.6]{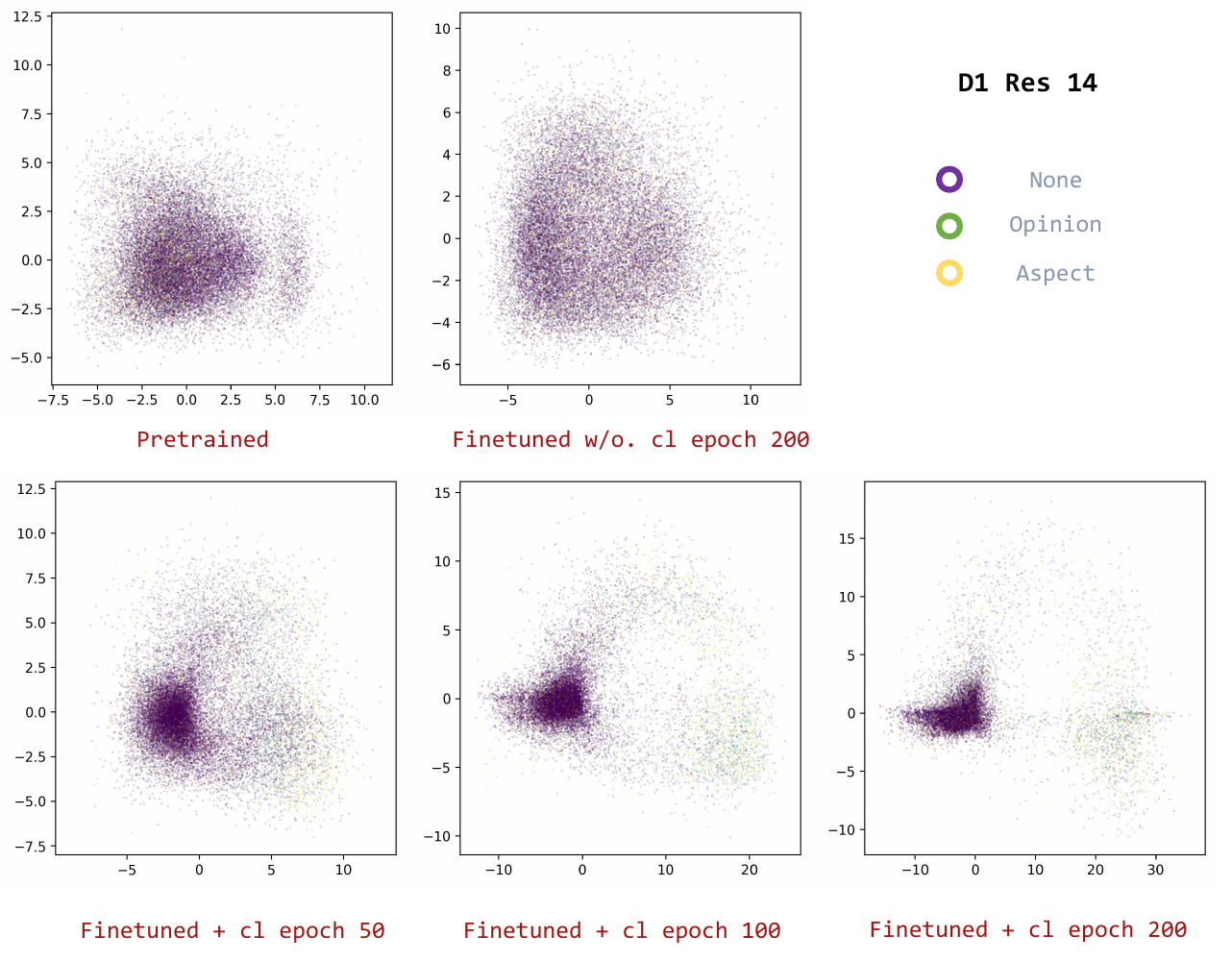} 
\caption{A plot of the hidden word representation, where the dimension is reduced to 2 for convenience of display. ``Pretrained'' means the model with official released version of weights. ``Finetuned'' means the model is a finetuned version on ASTE task for certain epochs. ``w/o. cl'' means the model is trained without contrastive learning loss. ``+ cl'' means the model is trained with contrastive learning. All the plotted results are from experiment carried on $\mathcal{D}_1$ 14Res. }
\label{fig.clrepre}
\end{center}
\end{figure*}


\subsection{Efficiency Analysis}

\begin{table}[htbp]
  \centering
  
  \scalebox{0.51}{
    \begin{tabular}{ccccccc}
    \toprule
    Model & Memory & Num Params & Epoch Time & Inf Time & F1(\%) & Device \\
    \midrule
    Span-ASTE & 3.173GB & - & 108s & - & 71.62 & Tesla v100 \\
    BDTF & 8.103GB & - & 135s & - & 74.73 & Tesla v100 \\
    GPT 3.5-Turbo & $>80GB^\natural$ & 175B$^\dagger$ & - & 0.83s & 49.48 & OpenAI API \\
    GPT 4 & $>80GB^\natural$ & 1760B$^\ddagger$ & - & 1.56s & 35.37 & OpenAI API \\
    \midrule
    Ours  & 7.11GB & 0.12B & 10s & 0.01s &  76.00    & 2080 Ti \\
    \bottomrule
    \end{tabular}%
      }
\caption{Efficiency Analysis, where $^\dagger$ is evaluated by \cite{gao2021sizes} and later confirmed by OpenAI \cite{GPT3Wiki64:online}, $^\ddagger$ is estimated by \cite{schreiner2023gpt4}, and $^\natural$ is reported by \cite{GPT3Wiki64:online}. }
  \label{tab:a1}%

\end{table}%


\begin{table}[htbp]
  \centering

  \scalebox{0.62}{
    \begin{tabular}{cccc}
    \toprule
    Method & Num Tags& Linguistic Features& Half/Full Matrix\\
    \midrule
    GTS & 6 & None & Half \\
    Double-encoder & 9 & None & Half \\
    EMC-GCN & 10 & 4 Groups & Full \\
    BDTF & $2\times2\times3$ & None & Half \\
    STAGE & $2\times2\times4$ & None & Half \\
    DGCNAP & 6 & POS-tagging & Half \\
    \midrule
    Ours  & $5$ & None & Full    \\
    \bottomrule
    \end{tabular}%
      }
        \caption{Tagging Scheme Comparisons}
  \label{tab:tagging}%
\end{table}%

Table \ref{tab:a1} shows an efficiency analysis, where our method is in significantly higher efficiency. 
While our approach demonstrates efficient management of memory and parameter utilization, surpassing other methods in terms of runtime performance, it is crucial to note the unsatisfactory performance of LLM-based methods, as demonstrated by Table \ref{tab-performance1}, Table \ref{tab-performance2} and Table \ref{gptExample}. This observation highlights that employing a general LLM with a large number of parameters does not yield desirable results in specific tasks like ASTE, even with the assistance of few-shot memorizing processes. Moreover, the incorporation of LLM introduces a significant computational resource overhead in general application scenarios. Although fine-tuning the parameters of LLM itself may offer some improvement, there exists a risk of collapsive forgetting.
Furthermore, Table \ref{tab:tagging} provides a comparative analysis of tagging schemes. Our method achieves its objective without the need for additional linguistic information and utilizes fewer tags.

\section{Related Work}

\subsection{ASTE Paradigms}

\citeauthor{peng2020knowing} (\citeyear{peng2020knowing}) proposes a pipeline method to bifurcate ASTE tasks into two stages, extracting \texttt{(Aspect, opinion)} pairs initially and predicting sentiment polarity subsequently. However, the error propagation hampers pipeline methods, rendering them vulnerable to end-to-end counterparts. 
End-to-end strategies, under the sequence-labeling approach, treat ASTE as a 1D ``\texttt{B-I-O}'' tagging scheme. ET \cite{xu2020position} introduces a position-aware tagging scheme with a conditional random field (CRF) module, effectively addressing span overlapping issues.
Recent advances in the end-to-end paradigm delicately grasp the peculiarity of ASTE tasks and come up with a proficient 2D table-filling tagging scheme. 
Other researches treat ASTE as generation problem, and develops seq2seq methods such as machine reading comprehension \cite{zhai2022mrc, mao2021joint, zou2024multi, chen2021semantic}. 

\subsection{Tagging Schemes}

\citeauthor{wu2020grid} (\citeyear{wu2020grid}) pioneer the adoption of a grid tagging scheme (GTS) for ASTE, yielding substantial performance gains. 
While subsequent research refines and enhances GTS, it is not devoid of drawbacks. Instances involving multi-word \texttt{Aspect}/\texttt{Opinion} constructs risk \textit{relation inconsistency} and \textit{boundary insensitivity} \cite{zhang2022boundary}. 
To overcome these, BDTF \cite{zhang2022boundary} designs a boundary-driven tagging scheme, effectively reducing boundary prediction errors. 
Alternative research augments GTS by integrating external semantic information as structured knowledge into their models. $\mathrm{S}^3\mathrm{E}^2$ \cite{chen2021semantic} retains the GTS tagging scheme while introducing novel semantic and syntactic enhancement modules between word embedding outputs and the tagging scheme. EMGCN \cite{chen2022enhanced} offers a distinct perspective, incorporating external knowledge from four aspects, namely, Part-of-Speech Combination, Syntactic Dependency Type, Tree-based Distance, and Relative Position Distance through an exogenous hard-encoding strategy. 
SyMux \cite{fei2022inheriting} contributes a unified tagging scheme capable of all ABSA subtasks synthesizing insights from incorporating GCN, syntax encoder, and representation multiplexing. 



\subsection{Contrastive Learning}



While contrastive learning has gained popularity in diverse NLP domains \cite{wu2020clear, giorgi2021declutr, gao2021simcse, zhang2021supporting}, its application to ASTE remains relatively unexplored. \citeauthor{ye2021contrastive} (\citeyear{ye2021contrastive}) adopts contrastive learning into triplet extraction in a generative fashion. \citeauthor{wang2022contrastive} (\citeyear{wang2022contrastive}) takes contrastive learning as a data augmentation approach. \citeauthor{yang2023pairing} (\citeyear{yang2023pairing}) proposed an enhancement approach in pairing 
with two separate encoders.

\section{Conclusion}

In this work, we have introduced an elegant and efficient framework for ASTE, achieving SOTA performance. Our approach is built upon two effective components: a new tagging scheme and a novel token-level contrastive learning implementation. The ablation study demonstrates the synergy between these components, reducing the need for complex model designs and external information enhancements. 

\section{Limitations \& Future Work}

Our work focused on ASTE problem for English, and can not ensure that our framework will work so well in other languages. However, our conventional 2D tagging method doesn't care much about grammar or other rules of a language. We believe that our framework will apply in other languages.  

Our study is purely based on the table-filling paradigm of ASTE approaches. In future work, it is worth exploring a combination between our method with other paradigms, such as seq2seq approaches. 

Finally, our framework may not be the best. It lefts potential to further investigate various classification head strategies.


\section{Ethics Statement}
In all our experiments, we employed datasets that are widely accepted and extensively referenced within the academic community. These datasets primarily focus on reviews of products and services on e-commerce platforms, which inherently possess a lower risk of containing offensive content. We have made every effort to scrutinize the data for potential biases against gender, race, and marginalized groups.

Despite our precautions, it is important to note that our model might still generate sentiment assessments that could be perceived as offensive, particularly if deployed in inappropriate contexts, such as evaluating statements related to ethical or moral issues. In such cases, we reserve the right to limit or modify the use of our technology to prevent misuse. 

\newpage

\bibliography{custom}


\newpage

\appendix

\section{Appendix}
\label{sec:Appendix}

\subsection*{Proof 1: }
\begin{adjustwidth}{0.9em}{0pt}
Let:
\begin{itemize}
    \item \( S \) be a sentence with \( n \) tokens.
    \item \( M \) be an \( n \times n \) tagging matrix for \( S \), where each entry \( M[i][j] \) can hold a label.
    \item \( T_k = (A_k, O_k, S_k) \) be a sentiment triplet consisting of an aspect term \( A_k \), an opinion term \( O_k \), and a sentiment \( S_k \).
\end{itemize}

\paragraph{Tagging Strategy}
If \( A_k \) starts at position \( i \) and \( O_k \) starts at position \( j \), then \( M[i][j] \) is tagged with a unique label \( L_k \) that encodes \( S_k \). This label \( L_k \) uniquely identifies the triplet \( T_k \), ensuring that no other entry \( M[i'][j'] \) with \( (i', j') \neq (i, j) \) carries the same label unless it refers to the same sentiment context.

\[
\text{Define } L_k = \text{"start of triplet"} T_k \text{ with sentiment } S_k
\]

\paragraph{Proof of One-to-One Mapping}
\begin{itemize}
    \item \textbf{Injectivity}: Each \( L_k \) uniquely identifies a triplet \( T_k \). If \( M[i][j] = M[i'][j'] = L_k \), then by definition, \( (i, j) = (i', j') \) and \( T_k \) is the same.
    \item \textbf{Surjectivity}: Each triplet \( T_k \) can be uniquely located and identified by its label \( L_k \) in matrix \( M \), where no two distinct triplets have the same label at the same matrix position.
\end{itemize}

\paragraph{Conclusion}
The tagging scheme ensures that each sentiment triplet \( T_k \) is uniquely mapped to a specific label in the matrix \( M \), and each label in \( M \) uniquely refers back to a specific triplet \( T_k \). This guarantees a one-to-one correspondence between the triplets and their tagging matrix representations, fulfilling the conditions required by Lemma 1 for an effective and efficient ASTE process.

\end{adjustwidth}

\subsection*{Proof 2: }
\begin{adjustwidth}{0.9em}{0pt}
For the ASTE task, considered as a 2D-labeling problem, it is necessary to ensure three fundamental goals are met:

\paragraph{Definitions}
\begin{itemize}
    \item \( S \) be a sentence with \( n \) tokens.
    \item \( M \) be an \( n \times n \) tagging matrix for \( S \), where each entry \( M[i][j] \) can hold a label indicating a component of a sentiment triplet.
    \item \( T_k = (A_k, O_k, S_k) \) be a sentiment triplet consisting of an aspect term \( A_k \), an opinion term \( O_k \), and a sentiment \( S_k \).
\end{itemize}

\paragraph{Goals}
\begin{enumerate}
    \item \textbf{Correct Identification of Pairs}: Ensure that each (Aspect, Opinion) pair is correctly identified in the tagging matrix \( M \).
    \item \textbf{Classification of Sentiment Polarity}: Accurately classify the sentiment polarity \( S_k \) for each (Aspect, Opinion) pair.
    \item \textbf{Avoidance of Boundary Errors}: Prevent boundary errors such as overlapping and confusion in the tagging matrix \( M \).
\end{enumerate}

\paragraph{Proof Using Contraposition}
\begin{enumerate}
    \item \textbf{Assuming Incorrect Identification}:
    Assume that some (Aspect, Opinion) pairs are incorrectly identified in \( M \). This would mean that there exists at least one pair \( (i, j) \) where \( M[i][j] \) does not represent the actual (Aspect, Opinion) relationship in \( S \). This misrepresentation leads to incorrect sentiment analysis results, which contradicts the requirement of the task to provide accurate sentiment analysis, thereby proving that our identification must be correct.

    \item \textbf{Assuming Incorrect Classification}:
    Assume the sentiment polarity \( S_k \) is incorrectly classified in \( M \). This would imply that the sentiment associated with an (Aspect, Opinion) pair is wrong, leading to a sentiment analysis that does not reflect the true sentiment of the text. Given that the primary goal of ASTE is to accurately identify sentiments, this assumption leads to a contradiction, thereby establishing that our classification must be accurate.

    \item \textbf{Assuming Existence of Boundary Errors}:
    Assume boundary errors such as overlaps or confusion occur in \( M \). Such errors would prevent the clear identification and classification of sentiment triplets, leading to incorrect or ambiguous extraction outcomes. This would undermine the integrity and usability of the ASTE process, contradicting the task's need for precise extraction mechanisms. Hence, we prove that boundary errors must be effectively managed.
\end{enumerate}

\paragraph{Conclusion}
The contraposition approach solidifies that the tagging strategy for ASTE in a 2D labeling framework successfully achieves the correct identification of pairs, accurate classification of sentiment, and effective management of boundary errors, as any failure in these aspects leads to contradictions with the task requirements.

\end{adjustwidth}

\begin{algorithm}[!ht]
\small

\caption{Workflow of our framework.}

\textbf{Modules}:









\textbf{Input}: 

\ \ Raw sentences: $\mathcal{S}_{|\mathcal{S}|}$; 

\ \ Ground truth triplets: $\mathcal{T}_{|\mathcal{T}|}^{gt}$ , where

\ \ \ $\mathcal{T}_k=(A_k, O_k, S_k)$, $k \in \{1,2,...,|\mathcal{T}|\}$;

\ \  classes of contrasted labels: $\mathcal{C}$.

\textbf{Output}: 



\ \ Predicted Triplets: $\mathcal{T}_{|\mathcal{T}|}^{pred}$;

\ \ Metric: $\mathit{Precision}, \mathit{Recall}, \mathit{F1}$.

\textbf{Algorithm}:\\
Repeat for $N$ epochs:

\begin{algorithmic}[1] 
\STATE Hidden word representation: \\ $\mathcal{H}_{|\mathcal{H}|} = \texttt{PLMsEncoder}(\mathcal{S}_{|\mathcal{S}|})$;
\STATE Tensor Operations: 

 $\mathcal{H}_{|\mathcal{H}|\times |\mathcal{H}|}=\texttt{expand}(\mathcal{H}_{|\mathcal{H}|})$, 

 $\mathcal{H}_{|\mathcal{H}|\times |\mathcal{H}|}^T=\mathcal{H}_{|\mathcal{H}|\times |\mathcal{H}|}.\texttt{transpose}()$;
\STATE Similarity matrix: 

$\mathbf{Sim}_{|\mathcal{H}|\times |\mathcal{H}|}=- (\mathcal{H}_{|\mathcal{H}|\times |\mathcal{H}|}-\mathcal{H}_{|\mathcal{H}|\times |\mathcal{H}|}^T) \circ (\mathcal{H}_{|\mathcal{H}|\times |\mathcal{H}|}-\mathcal{H}_{|\mathcal{H}|\times |\mathcal{H}|}^T) $, where \  $\mathbf{Sim}_{i, j} = -\dblvert{\mathcal{H}_i - \mathcal{H}_j}^2$, and $\circ$ denotes the Hadamard product;
\STATE Contrastive Mask matrix: $\mathbf{M}_{|\mathcal{H}|\times |\mathcal{H}|}$, where \  $\mathbf{M}_{i, j} = 1$\ if 
$\mathcal{H}_i, mathcal{H}_j\in \mathcal{C}_p, p\in{1, 2 , 3}$ 
else $-1$;

\STATE Contrastive loss: 

$\mathcal{L}_{contrastive}=\newline\sum_{i=1}^{|\mathcal{H}|}\sum_{j=1}^{|\mathcal{H}|}\big(\mathbf{Sim_{|\mathcal{H}|\times|\mathcal{H}|}} \circ \mathbf{M_{|\mathcal{H}|\times|\mathcal{H}|}}\big)_{i, j};$

\STATE Predicted tagging matrix: \\
$\mathbf{Tag}_{|\mathcal{H}|\times |\mathcal{H}|}^{pred}=\texttt{ClsHead}(\mathcal{H}_{|\mathcal{H}|\times |\mathcal{H}|}, \mathcal{H}_{|\mathcal{H}|\times |\mathcal{H}|}^T)$;
\STATE Focal loss: \\
$\mathcal{L}_{focal} = \texttt{FocalLoss}(\mathbf{Tag}_{|\mathcal{H}|\times |\mathcal{H}|}^{pred}, \mathbf{Tag}_{|\mathcal{H}|\times |\mathcal{H}|}^{gt})$;
\STATE Weighted Loss: $\mathcal{L}=\mathcal{L}_{focal} + \alpha \mathcal{L}_{contrastive}$.
\STATE Backward propagation.

\end{algorithmic}

Predicted triplets: 

\ \ $\mathcal{T}_{|\mathcal{T}|}^{pred}=\texttt{TaggingDecoder}(\mathbf{Tag}_{|\mathcal{H}|\times |\mathcal{H}|}^{pred})$

Metric:

\ \ $\mathit{Precision}, \mathit{Recall}, \mathit{F1} = \texttt{Metric}( \mathcal{T}_{|\mathcal{T}|}^{pred}, \mathcal{T}_{|\mathcal{T}|}^{gt})$

\label{alg1}
\end{algorithm}

\begin{table*}[ht]
  \centering
  \scalebox{0.8}{
    \begin{tabular}{cccccccccc}
    \hline
    \toprule
    \multicolumn{3}{c}{\textbf{Datasets}} & \textbf{\#S} & \textbf{\#A} & \textbf{\#O} & \textbf{\#S1} & \textbf{\#S2 } & \textbf{\#S3} & \textbf{\#T} \\
    \hline
    \midrule
    
    \multirow{6}[4]{*}{\textbf{14Res}} & \multirow{3}[2]{*}{$\mathcal{D}_1$} & \textbf{Train} & 1259  & 1008  & 849   & 1456  & 164   & 446   & 2066 \\
          &       & \textbf{Dev} & 315   & 358   & 321   & 352   & 44    & 93    & 489 \\
          &       & \textbf{Test} & 493   & 591   & 433   & 651   & 59    & 141   & 851 \\
\cmidrule{2-10}          & \multirow{3}[2]{*}{$\mathcal{D}_2$} & \textbf{Train} & 1266  & 986   & 844   & 1692  & 166   & 480   & 2338 \\
          &       & \textbf{Dev} & 310   & 396   & 307   & 404   & 54    & 119   & 577 \\
          &       & \textbf{Test} & 492   & 579   & 437   & 773   & 66    & 155   & 994 \\
    \midrule
    \multirow{6}[4]{*}{\textbf{14Lap}} & \multirow{3}[2]{*}{$\mathcal{D}_1$} & \textbf{Train} & 899   & 731   & 693   & 691   & 107   & 466   & 1264 \\
          &       & \textbf{Dev} & 225   & 303   & 237   & 173   & 42    & 118   & 333 \\
          &       & \textbf{Test} & 332   & 411   & 330   & 305   & 62    & 101   & 468 \\
\cmidrule{2-10}          & \multirow{3}[2]{*}{$\mathcal{D}_2$} & \textbf{Train} & 906   & 733   & 695   & 817   & 126   & 517   & 1460 \\
          &       & \textbf{Dev} & 219   & 268   & 237   & 169   & 36    & 141   & 346 \\
          &       & \textbf{Test} & 328   & 400   & 329   & 364   & 63    & 116   & 543 \\
    \midrule
    \multirow{6}[4]{*}{\textbf{15Res}} & \multirow{3}[2]{*}{$\mathcal{D}_1$} & \textbf{Train} & 603   & 585   & 485   & 668   & 24    & 179   & 871 \\
          &       & \textbf{Dev} & 151   & 182   & 161   & 156   & 8     & 41    & 205 \\
          &       & \textbf{Test} & 325   & 353   & 307   & 293   & 19    & 124   & 436 \\
\cmidrule{2-10}          & \multirow{3}[2]{*}{$\mathcal{D}_2$} & \textbf{Train} & 605   & 582   & 462   & 783   & 25    & 205   & 1013 \\
          &       & \textbf{Dev} & 148   & 191   & 183   & 185   & 11    & 53    & 249 \\
          &       & \textbf{Test} & 322   & 347   & 310   & 317   & 25    & 143   & 485 \\
    \midrule
    \multirow{6}[4]{*}{\textbf{16Res}} & \multirow{3}[2]{*}{$\mathcal{D}_1$} & \textbf{Train} & 863   & 775   & 602   & 890   & 43    & 280   & 1213 \\
          &       & \textbf{Dev} & 216   & 270   & 237   & 224   & 8     & 66    & 298 \\
          &       & \textbf{Test} & 328   & 342   & 282   & 360   & 25    & 72    & 457 \\
\cmidrule{2-10}          & \multirow{3}[2]{*}{$\mathcal{D}_2$} & \textbf{Train} & 857   & 759   & 623   & 1015  & 50    & 329   & 1394 \\
          &       & \textbf{Dev} & 210   & 251   & 221   & 252   & 11    & 76    & 339 \\
          &       & \textbf{Test} & 326   & 338   & 282   & 407   & 29    & 78    & 514 \\
    \bottomrule
    \hline
    \end{tabular}%
    
    }
  \caption{Statistic information of our two experiment datasets: 
  ``\#S'', ``\#T'', ``\#A'', and ``\#O'' denote the numbers of ``Sentences'', ``Triplets'', ``Aspects'', and ``Opinions''; ``\#S1'', ``\#S2'', \#S3'' denote the numbers of sentiments ``Positive'', ``Neutral'' and ``Negative'', respectively.}
  \label{tab:data1}%
\end{table*}%

\begin{table*}[htbp]
  \centering
  \scalebox{0.68}{
    \begin{tabular}{cccccccccccccccc}
    \hline
    \toprule
    \multirow{2}[4]{*}{\textbf{Methods}} & \multicolumn{3}{c}{\textbf{14Res}} &       & \multicolumn{3}{c}{\textbf{14Lap}} &       & \multicolumn{3}{c}{\textbf{15Res}} &       & \multicolumn{3}{c}{\textbf{16Res}} \\
\cmidrule{2-4}\cmidrule{6-8}\cmidrule{10-12}\cmidrule{14-16}          & \textbf{AE} & \textbf{OE} & \textbf{AOPE} &       & \textbf{AE} & \textbf{OE} & \textbf{AOPE} &       & \textbf{AE} & \textbf{OE} & \textbf{AOPE} &       & \textbf{AE} & \textbf{OE} & \textbf{AOPE} \\
    \midrule
    \midrule
    CMLA  & 81.22 & 83.07 & 48.95 &       & 78.68 & 77.95 & 44.10 &       & 76.03 & 74.67 & 44.60 &       & 74.20 & 72.20 & 50.00 \\
    RINANTE & 81.34 & 83.33 & 46.29 &       & 77.13 & 75.34 & 29.70 &       & 73.38 & 75.40 & 35.40 &       & 72.82 & 70.45 & 30.70 \\
    Li-unified & 81.62 & 85.26 & 55.34 &       & 78.54 & 77.55 & 52.56 &       & 74.65 & 74.25 & 56.85 &       & 73.36 & 73.87 & 53.75 \\
    GTS   & 83.82 & 85.04 & 75.53 &       & 79.52 & 78.61 & 65.67 &       & 78.22 & 79.31 & 67.53 &       & 75.80 & 76.38 & 74.62 \\
    Dual-MRC & \textbf{86.60} & 86.22 & 77.68 &       & 80.44 & 79.90 & 63.37 &       & 75.08 & 77.52 & 64.97 &       & 76.87 & 77.90 & 75.71 \\
    \midrule
    \textbf{ContrASTE (Ours)} & 86.55 & \textbf{87.04} & \textbf{79.60} &       & \textbf{82.62} & \textbf{83.41} & \textbf{73.23} &       & \textbf{86.53} & \textbf{83.05} & \textbf{73.87} &       & \textbf{85.48} & \textbf{87.06} & \textbf{76.29} \\
    $\Delta$F1     & -0.05 & 0.82  & 1.92  &       & 2.18  & 3.51  & 7.56  &       & 8.31 & 3.74  & 6.34  &       & 8.61  & 9.16  & 0.58 \\

    \bottomrule
    \hline
    \end{tabular}%
  }
  \caption{F1-score performance on other ABSA tasks: AE, OE, and AOPE. The test is implemented on $\mathcal{D}_1$. Results of other models are retrieved from \cite{fei2022inheriting}. }
  \label{ABSA}%
\end{table*}%

\begin{table*}[ht]
\centering
  \scalebox{0.68}{
    \begin{tabular}{cccccccccccccccc}
    \hline
    \toprule
    \multirow{2}[4]{*}{\textbf{Methods}} & \multicolumn{3}{c}{\textbf{14Res}} &       & \multicolumn{3}{c}{\textbf{14Lap}} &       & \multicolumn{3}{c}{\textbf{15Res}} &       & \multicolumn{3}{c}{\textbf{16Res}} \\
\cmidrule{2-4}\cmidrule{6-8}\cmidrule{10-12}\cmidrule{14-16}          & \multicolumn{1}{c}{P} & R     & F1    &       & \multicolumn{1}{c}{P} & \multicolumn{1}{c}{R} & \multicolumn{1}{c}{F1} &       & \multicolumn{1}{c}{P} & R     & F1    &       & \multicolumn{1}{c}{P} & \multicolumn{1}{c}{R} & \multicolumn{1}{c}{F1} \\
    \midrule
    \midrule
    \textbf{Pipeline} &       &       &       &       &       &       &       &       &       &       &       &       &       &       &  \\
    OTE-MTL \textrm{\cite{zhang2020multi}} & -     & -     & 45.05 &       & -     & -     & 59.67 &       & -     & -     & 48.97 &       & -     & -     & 55.83 \\
    $\textrm{Li-unified-R+PD}^\sharp$\ \textrm{\cite{peng2020knowing}} & 41.44     & 68.79     & 51.68 &       & 42.25     & 42.78     & 42.47 &       & 43.34     & 50.73     & 46.69 &       & 38.19     & 53.47     & 44.51 \\
    \textrm{RI-NANTE+} \textrm{\cite{dai2019neural}} & 31.42 & 39.38 & 34.95 &       & 21.71 & 18.66 & 20.07 &       & 29.88 & 30.06 & 29.97 &       & 25.68 & 22.30 & 23.87 \\
    $\textrm{CMLA+C-GCN}^\flat$\ \textrm{\cite{wang2017coupled}} & 72.22 & 56.35 & 63.17 &       & 60.69 & 47.25 & 53.03 &       & 64.31 & 49.41 & 55.76 &       & 66.61 & 59.23 & 62.70 \\
    Two-satge$^\natural$\ \textrm{\cite{peng2020knowing}} & 58.89 & 60.41 & 59.64 &       & 48.62 & 45.52 & 47.02 &       & 51.7  & 46.04 & 48.71 &       & 59.25 & 58.09 & 59.67 \\
    \midrule
    \textbf{Sequence-tagging} &       &       &       &       &       &       &       &       &       &       &       &       &       &       &  \\
    Span-BART \textrm{\cite{yan2021unified}} & -     & -     & 72.46 &       & -     & -     & 57.59 &       & -     & -     & 60.10 &       & -     & -     & 69.98 \\
    JET \textrm{\cite{xu2020position}}   & 67.97 & 60.32 & 63.92 &       & 58.47 & 43.67 & 50.00 &       & 58.35 & 51.43 & 54.67 &       & 64.77 & 61.29 & 62.98 \\
    \midrule
    \textbf{MRC based} &       &       &       &       &       &       &       &       &       &       &       &       &       &       &  \\
    BMRC$^\dagger$\ \textrm{\cite{chen2021bidirectional}}  & 71.32 & 70.09 & 70.69 &       & 65.12 & 54.41 & 59.27 &       & 63.71 & 58.63 & 61.05 &       & 67.74 & 68.56 & 68.13 \\
    COM-MRC \textrm{\cite{zhai2022mrc}} & \underline{76.45} & 69.67 & 72.89 &       & 64.73 & 56.09 & 60.09 &       & 68.50 & 59.74 & 63.65 &       & \underline{72.80} & 70.85 & 71.79 \\
    \midrule
    \textbf{Table-filling} &       &       &       &       &       &       &       &       &       &       &       &       &       &       &  \\
    $\mathrm{S}^3\mathrm{E}^2$ \textrm{\cite{chen2021semantic}} & 69.08 & 64.55 & 66.74 &       & 59.43 & 46.23 & 52.01 &       & 61.06 & 56.44 & 58.66 &       & 71.08 & 63.13 & 66.87 \\
    GTS \textrm{\cite{wu2020grid}}   & 70.92 & 69.49 & 70.20 &       & 57.52 & 51.92 & 54.58 &       & 59.29 & 58.07 & 58.67 &       & 68.58 & 66.60 & 67.58 \\
    EMC-GCN \textrm{\cite{chen2022enhanced}} & 71.85 & 72.12 & 71.78 &       & 61.46 & \underline{55.56} & 58.32 &       & 59.89 & 61.05 & 60.38 &       & 65.08 & 71.66 & 68.18 \\
    BDTF \textrm{\cite{zhang2022boundary}} & \textbf{76.71} & \underline{74.01} & \underline{75.33} &       & \textbf{68.30}  & 55.10  & \underline{60.99} &       & \textbf{66.95} & \textbf{65.05} & \textbf{65.97} &       & \textbf{73.43} & \underline{73.64} & \textbf{73.51} \\
    DGCNAP \textrm{\cite{li2023dual}} & 71.83 & 68.77 & 70.26 &       & 66.46 & 54.34 & 58.74 &       & 62.03 & 57.18  & 59.49 &       & 69.39 & 72.20 & 70.77 \\
    \midrule

    \textbf{LLM-based} &       &       &       &       &       &       &       &       &       &       &       &       &       &       &  \\
    GPT 3.5 zero-shot  & 39.21     & 56.17     & 46.18 &       & 26.21     & 40.69     & 31.88 &       & 31.21     & 52.75     & 39.21 &       & 35.28     & 59.64    & 44.34 \\
    $\textrm{GPT 3.5 few-shots}$\  & 44.73     & 58.87     & 50.84 &       & 29.63     & 37.69     & 33.18 &       & 37.27     & 56.42     & 44.89 &       & 43.15     & 60.75     & 50.46 \\
    \textrm{GPT 4 zero-shot}  & 27.34 & 37.13 & 31.49 &       & 16.50 & 24.41 & 19.69 &       & 25.60 & 39.22 & 30.98 &       & 28.39 & 43.64 & 35.79 \\
    $\textrm{GPT 4 few-shots}$\  & 41.48 & 52.06 & 46.17 &       & 28.79 & 39.83 & 33.42 &       & 38.04 & 58.02 & 45.96 &       & 40.89 & 62.50 & 49.44 \\
    \midrule
    
    \textbf{Ours} &       &       &       &       &       &       &       &       &       &       &       &       &       &       &  \\
      ContrASTE    & 75.87 & \textbf{76.12} & \textbf{76.00} &       & \underline{67.45} & \textbf{61.01} & \textbf{64.07} &       & \underline{66.84} & \underline{64.08} & \underline{65.43} &       & 69.38 & \textbf{74.40} & \underline{71.80} \\
    \bottomrule
    \hline
    \end{tabular}
}
    \caption{Experimental results on $\mathcal{D}_1$ \cite{wu2020grid}. The best results are highlighted in bold,  while the second best results are underlined.
}
\label{tab-performance1}
\end{table*}

\begin{table*}[htbp]
\centering
  \scalebox{0.65}{
    \begin{tabular}{lp{40em}}
    \hline
    \toprule
    \textbf{Methods} &  \textbf{Brief Introduction}  \\
    \midrule
    \midrule
    \textbf{Pipeline} &    \\
    OTE-MTL \textrm{\cite{zhang2020multi}} & It proposes a multi-task learning framework including two parts: aspect and opinion tagging, along with word-level sentiment dependency parsing. This approach simultaneously extracts aspect and opinion terms while parsing sentiment dependencies using a biaffine scorer. Additionally, it employs triplet decoding based on the aforementioned outputs during inference to facilitate triplet extraction. \\
    \textrm{Li-unified-R+PD}\ \textrm{\cite{peng2020knowing}} &   It proposes an unified tagging scheme, Li-unified-R, to assist target boundary detection. Two stacked LSTMs are employed to complete aspect-based sentiment prediction and the sequence labeling. \\
    \textrm{CMLA+C-GCN}\ \textrm{\cite{wang2017coupled}} & It facilitates triplet extraction by modelling the interaction between the aspects and opinions.  \\
    Two-satge\ \textrm{\cite{peng2020knowing}} & It decomposes triplet extraction to two stages: 1)  predicting unified aspect-sentiment and opinion tags; and 2) pairing the two results from stage one. \\
    \textrm{RI-NANTE+} \textrm{\cite{dai2019neural}} & It adopts the same sentiment triplets extracting method as that of \textrm{CMLA+}, but it incorporates a novel LSTM-CRF mechanism and fusion rules to capture word dependencies within sentences.\\

    \midrule
    \textbf{Sequence-tagging} &       \\
    Span-BART \textrm{\cite{yan2021unified}} & It redefines triplet extraction within an end-to-end framework by utilizing a sequence composed of pointer and sentiment class indexes. This is achieved by leveraging the pretrained sequence-to-sequence model BART to address ASTE. \\
    JET \textrm{\cite{xu2020position}}   &  It extracts triplets jointly by designing a position-aware sequence-tagging scheme to extract the triplets and capturing the rich interactions among the elements.  \\
    \midrule
    \textbf{Seq2seq} &    \\
    Dual-MRC \textrm{\cite{mao2021joint}} & It proposes a solution for ASTE by jointly training two BERT-MRC models with parameters sharing. \\
    BMRC \textrm{\cite{chen2021bidirectional}} & It introduces a bidirectional MRC (BMRC) framework for ASTE, employing three query types: non-restrictive extraction queries, restrictive extraction queries, and sentiment classification queries. The framework synergistically leverages two directions, one for sequential recognition of aspect-opinion-sentiment and the other for sequential recognition of opinion-aspects-sentiment expressions. \\
    \midrule
    \textbf{Table-filling} &   \\
    GTS \textrm{\cite{wu2020grid}}   & It proposes a novel 2D tagging scheme to address ASTE in an end-to-end fashion only with one unified grid tagging task. 
    It also devises an effective inference strategy on GTS that utilizes mutual indication between different opinion factors to achieve more accurate extraction. \\
    Double-encoder \textrm{\cite{jing2021seeking}} & It proposes a dual-encoder model that capitalizes on encoder sharing while emphasizing differences to enhance effectiveness. 
    One of the encoders, referred to as the pair encoder, specifically concentrates on candidate aspect-opinion pair classification, while the original encoder retains its focus on sequence labeling. \\
    $\mathrm{S}^3\mathrm{E}^2$ \textrm{\cite{chen2021semantic}} &  It represents the semantic and syntactic relationships between word pairs, employs GNNs for encoding, and applies a more efficient inference strategy. \\
    EMC-GCN \textrm{\cite{chen2022enhanced}} & It employs a biaffine attention module to embed ten types of relations within sentences, transforming the sentence into a multi-channel graph while incorporating various linguistic features to enhance performance. 
    Additionally, the method introduces an effective strategy for refining word-pair representations, aiding in the determination of whether word pairs are a match or not.\\
    \midrule
    \textbf{LLM-based} &    \\
    zero-shot & Performing aspect-based sentiment analysis using an LLM. The specific method involves inputting a prompted sentence and directly outputting the corresponding [A, O, S] triplets. An example of the text given to the LLM, with the prompt added, is as follows: \textcolor{cyan}{"Perform aspect-based sentiment analysis on the provided text and return triplets as [Aspect, Opinion, Sentiment]. You only need to provide the triplets, no additional explanations are required. The provided text: \{sentence\}"}\\ 

    few-shots & Building upon the zero-shot method, a small number of examples from the training set are added to the prompted sentence: \textcolor{cyan}{"Perform aspect-based sentiment analysis on the provided text and return triplets as [Aspect, Opinion, Sentiment]. For example: input: \{train sentence\} output: \{train triplets\}, ... (some other examples). You only need to provide the triplets, no additional explanations are required. The provided text: \{sentence\}"}\\
    \bottomrule
    \hline
    \end{tabular}%
    }
\caption{Baselines methods with brief introduction.}
\label{baseline}
\end{table*}

\begin{sidewaystable*}[htbp]
\centering
\tiny
 \renewcommand{\arraystretch}{1.5}
\rotatebox{0}{
\begin{tabular}{|p{5cm}|p{3cm}|p{3cm}|p{3cm}|p{3cm}|p{3cm}|p{3cm}|}
\hline
\multirow{2}{*}{Sentence} & \multirow{2}{*}{Ground Truth} & \multicolumn{4}{c|}{Predictions} \\ \cline{3-6} 
                     &                      & GPT-3.5 zero-shot & GPT-3.5 few-shots & GPT-4 zero-shot & GPT-4 few-shots \\ \hline \hline
It is a cozy place to go with a couple of friends. & [place, cozy, positive] & [place, cozy, positive],\newline[friends, couple of, positive] & [place, cozy, positive] & [place, cozy, positive]\newline[going with friends, a couple of, positive] & [place, cozy, positive] \\ \hline
This place has the best sushi in the city. & [sushi, best, positive] & [sushi, best, positive],\newline[place, best, positive], [city, best, positive] & [sushi, best, positive] & [sushi, the best, positive],\newline[place, This, neutral],\newline[city, in the, neutral] & [sushi, best, positive] \\ \hline
Disappointingly, their wonderful Saketini has been taken off the bar menu. & [Saketini, wonderful, positive],\newline[bar menu, Disappointingly, negative] & [Saketini, wonderful, positive],\newline[bar menu, taken off, negative] & [Saketini, taken off, negative] & [Saketini, has been taken off the bar menu, negative],\newline[bar menu, wonderful, positive] & [Saketini, wonderful, positive],\newline[bar menu, taken off, negative] \\ \hline
I went in one day asking for a table for a group and was greeted by a very rude hostess. & [hostess, rude, negative] & [hostess, rude, negative] & [hostess, rude, negative] & [table for a group, asking, neutral],\newline[hostess, very rude, negative] & [table for a group, asking, neutral],\newline[hostess, very rude, negative] \\ \hline
But make sure you have enough room on your credit card as the bill will leave a big dent in your wallet. & [bill, big, negative] & [credit card, enough room, positive],\newline[bill, big dent, negative],\newline[wallet, big dent, negative] & [credit card, enough room, negative],\newline[bill, big dent, negative],\newline[wallet, big dent, negative] & [room on your credit card, enough, positive],\newline[bill, will leave a big dent in your wallet, negative] & [bill, big dent, negative] \\ \hline
\end{tabular}
}

\caption{In summary, there are several challenges observed in the performance of GPT models concerning triplets. Firstly, there is a prominent issue of "hard" matching, where GPT models tend to introduce additional modifiers or adverbs in the opinion component, leading to a lack of exact correspondence. Secondly, during zero-shot inference, GPT models tend to generate multiple predicted triplets, resulting in decreased precision. This behavior particularly hampers the precision of the model's predictions. Thirdly, inconsistencies arise in handling triplets involving structures such as [A, O1 and O2, S] and [A, O1, S], [A, O2, S]. This inconsistency is challenging to mitigate due to its dependence on annotation practices and conventions.
Upon closer examination, the issues observed do not appear to be as pronounced as indicated by the evaluation metrics. Rather, they often manifest as cases where the general idea is correctly captured, but the precise format or phrasing does not align perfectly. Notably, the performance of GPT-4 deteriorates due to its occasional tendency to not merely "extract" fragments from sentences but to generate its own summarizations. Consequently, evaluating against triplets that originate solely from annotated sentences poses a challenge in achieving alignment. Furthermore, GPT-4 exhibits a proclivity for extracting longer sequences of words as aspects or opinions, while GPT-3.5 tends to produce shorter sequences that better conform to typical annotation scenarios.}

\label{gptExample}
\end{sidewaystable*}

\end{document}